\documentclass[smallextended]{svjour3}       
\smartqed  
\usepackage{amsmath}
\usepackage{graphicx}
\usepackage[utf8]{inputenc}

\usepackage[hyphens]{url} 
\usepackage{hyperref}

\usepackage{longtable,booktabs,array}
\usepackage{calc} 
\usepackage{etoolbox}
\makeatletter
\patchcmd\longtable{\par}{\if@noskipsec\mbox{}\fi\par}{}{}
\makeatother
\IfFileExists{footnotehyper.sty}{\usepackage{footnotehyper}}{\usepackage{footnote}}
\makesavenoteenv{longtable}

\newlength{\cslhangindent}
\newlength{\csllabelwidth}
\newlength{\cslentryspacingunit} 
  {}%
  {\par}
\newenvironment{CSLReferences}[2] 
 {
  \setlength{\parindent}{0pt}
  \ifodd #1
  \let\oldpar\par
  \def\par{\hangindent=\cslhangindent\oldpar}
  \fi
  \setlength{\parskip}{#2\cslentryspacingunit}
 }%
 {}
\usepackage{calc}

\usepackage{algorithm}
\usepackage{array}
\usepackage[noend]{algpseudocode}
\usepackage{amsmath}
\usepackage{amssymb}
\usepackage{amsfonts}
\usepackage{mathtools}
\usepackage{bm}
\usepackage{float}
\usepackage{tikz}
\usepackage{threeparttable}
\usetikzlibrary{trees,arrows, chains, fit, positioning, calc, shapes, shadows}
\usepackage{pdflscape}
\usepackage[T1]{fontenc}
\usepackage{orcidlink}
\usepackage{booktabs}
\usepackage{longtable}
\usepackage{array}
\usepackage{multirow}
\usepackage{wrapfig}
\usepackage{float}
\usepackage{colortbl}
\usepackage{pdflscape}
\usepackage{tabu}
\usepackage{threeparttable}
\usepackage{threeparttablex}
\usepackage[normalem]{ulem}
\usepackage{makecell}
\usepackage{xcolor}
\begin{document}

\title{Regularized target encoding outperforms traditional methods in supervised machine learning with high cardinality features }

    \titlerunning{Encoding high cardinality features in supervised ML}

\author{  Florian Pargent \orcidlink{0000-0002-2388-553X} \and  Florian Pfisterer \orcidlink{0000-0001-8867-762X} \and  Janek Thomas \orcidlink{0000-0003-4511-6245} \and  Bernd Bischl \orcidlink{0000-0001-6002-6980} \and  }

\institute{
        Florian Pargent \orcidlink{0000-0002-2388-553X} \at
     Department of Psychology, Psychological Methods and Assessment, LMU Munich, Leopoldstraße 13, 80802 Munich, Germany \\
     \email{\href{mailto:florian.pargent@psy.lmu.de}{\nolinkurl{florian.pargent@psy.lmu.de}}}  
    \and
        Florian Pfisterer \orcidlink{0000-0001-8867-762X} \at
     Department of Statistics, Statistical Learning and Data Science, LMU Munich, Ludwigstraße 33, 80539 Munich, Germany \\
    \and
        Janek Thomas \orcidlink{0000-0003-4511-6245} \at
     Department of Statistics, Statistical Learning and Data Science, LMU Munich, Ludwigstraße 33, 80539 Munich, Germany \\
    \and
        Bernd Bischl \orcidlink{0000-0001-6002-6980} \at
     Department of Statistics, Statistical Learning and Data Science, LMU Munich, Ludwigstraße 33, 80539 Munich, Germany \\
    \and
    }

\date{\textbf{This preprint has since been published in Computational Statistics:} \newline Pargent, F., Pfisterer, F., Thomas, J., \& Bischl, B. Regularized target encoding outperforms traditional methods in supervised machine learning with high cardinality features. \textit{Computational Statistics} (2022). https://doi.org/10.1007/s00180-022-01207-6 \newline \textbf{Link to copy of record:} https://link.springer.com/article/10.1007/s00180-022-01207-6}

\maketitle

\begin{abstract}
Since most machine learning (ML) algorithms are designed for numerical inputs, efficiently encoding categorical variables is a crucial aspect in data analysis. A common problem are high cardinality features, i.e.~unordered categorical predictor variables with a high number of levels. We study techniques that yield numeric representations of categorical variables which can then be used in subsequent ML applications. We focus on the impact of these techniques on a subsequent algorithm's predictive performance, and -- if possible -- derive best practices on when to use which technique. We conducted a large-scale benchmark experiment, where we compared different encoding strategies together with five ML algorithms (lasso, random forest, gradient boosting, k-nearest neighbors, support vector machine) using datasets from regression, binary- and multiclass- classification settings. In our study, regularized versions of target encoding (i.e.~using target predictions based on the feature levels in the training set as a new numerical feature) consistently provided the best results. Traditionally widely used encodings that make unreasonable assumptions to map levels to integers (e.g.~integer encoding) or to reduce the number of levels (possibly based on target information, e.g.~leaf encoding) before creating binary indicator variables (one-hot or dummy encoding) were not as effective in comparison.
\\

\end{abstract}

\def\spacingset#1{\renewcommand{\baselinestretch}%
{#1}\small\normalsize} \spacingset{1}

\hypertarget{introduction}{%
\section{Introduction}\label{introduction}}

While increasing sample size is usually considered the most important step to improve the predictive performance of a machine learning (ML) model, using effective feature engineering comes as a close second. One remaining challenge is how to handle high cardinality features -- categorical predictor variables with a high number of different levels but without any natural ordering.
While categorical variables with only a small number of possible levels can often be efficiently dealt with using standard techniques such as one-hot encoding, this approach becomes inefficient as the number of levels increases.
Despite this inefficiency, simpler strategies are often favored in practice because other methods are either not known, implementations are missing or because of a lack of trust due to missing validation studies.
Although domain knowledge can sometimes be used to reduce the number of theoretically relevant levels, finding strategies that work well on a large variety of problems is highly important for many applications as well as in automated ML (Thomas, Coors, and Bischl 2018; Thornton et al. 2013; Feurer et al. 2015). Optimally, strategies should be model-agnostic because benchmarking encoding methods together with ML algorithms from different classes is often necessary for applications.
While a variety of strategies exist, there are very few benchmarks that can be used to decide which technique is expected to yield good predictive performance.
Furthermore, there has recently been increasing attention on scientific benchmark studies that compare different methods to provide a clearer picture in light of a large number of methods available to practitioners (Fernández-Delgado et al. 2014; Bommert et al. 2020), as they can provide at least partial answers to such questions.
The goal of this study is to provide an overview of existing approaches for encoding categorical predictor variables and to study their effect on a model's predictive performance.
Following calls in the computational statistics community for neutral benchmark studies (Boulesteix et al. 2017), which do not introduce a new method, thus reducing the risk of cherry picking methods (Dehghani et al. 2021) and reporting over-optimistic performance (Nießl et al., n.d.), we present a carefully designed experimental setting to discern the effect of encoding strategies and their interaction with different ML algorithms.

\hypertarget{notation}{%
\subsection{Notation}\label{notation}}

We consider the classical setting of supervised learning from an \(i.i.d.\) tabular dataset \(\mathcal{D}\) of size \(N\) sampled from a joint distribution \(\mathcal{P}(\bm{x}, y)\) of a set of features \(\bm{x}\) and an associated target variable \(y\).
Here, \(\bm{x}\) consists of a mix of numeric (real-valued or integer-valued) features and categorical features, the latter of which we seek to transform feature-wise to numeric features using a categorical encoding technique.
Let \(x\) be a single unordered categorical feature from a feature space \(\mathcal{X}\) with cardinality \(card(\mathcal{X}) \leq card(\mathbb{N})\).
It holds either \(y \in \mathbb{R}\) (regression), \(y \in \mathcal{C}\) from a finite class space \(\mathcal{C} = \{c_1, ..., c_C\}\) with \(C = 2\) (binary classification) or \(C > 2\) (multiclass classification). We always assume to observe all \(C\) classes in our training sample, however we might only observe a subset \(\mathcal{L}^{train} \subseteq \mathcal{X}\) of a feature's available \(L\) levels, \(\mathcal{L}^{train} = \{l_1, ..., l_L\}\) for categorical features. We denote the observed frequency of class \(c\) in the training set with \(N_c\) and the observed frequency of a level \(l\) in the training set with \(N_l\).
We investigate categorical encoding techniques to transform each nominal feature \(x^{train}\) into numerical features \(\hat{x}^{train}\) which are then used for training. If clear from the context, we use \(\hat{x}_l\) as the encoded value for an observation with level \(l\). Although datasets might contain multiple high cardinality features, we encode each feature separately but with the same strategy.

\hypertarget{related-work}{%
\subsection{Related Work}\label{related-work}}

We broadly categorize feature encoding techniques into \emph{target-agnostic} methods and \emph{target-based} methods (Micci-Barreca 2001).
Figure \ref{fig:overview} contains a taxonomy of our considered encoding methods. \emph{Target-agnostic} methods do not rely on any information about the target variable and can therefore also be used in unsupervised settings.
Simple strategies from this domain e.g.~one-hot or dummy encoding are widely used -- in the scientific literature (Kuhn and Johnson 2019; Hancock and Khoshgoftaar 2020) but also on Kaggle\footnote{\url{https://www.kaggle.com}} to embed variables for classical ML algorithms as well as (deep) neural networks.
Such indicator methods map each level of a categorical variable to a set of dichotomous features encoding the presence or absence of a particular level.
An obvious drawback of indicator encoding is that it adds one additional feature per level of a categorical variable.
When indicator encoding leads to an unreasonable number of features, levels are often mapped to integer values with random order (integer encoding). Alternatively, the ``hashing trick'' (Weinberger et al. 2009) can be used to randomly collapse feature levels into a smaller number of indicator variables (Kuhn and Johnson 2019), or levels can be encoded by using the observed frequency of a given level in the dataset (frequency encoding).

\tikzset{
  basic/.style  = {draw, text width=4cm, drop shadow, font=\sffamily \scriptsize, rectangle},
  root/.style   = {basic, rounded corners=2pt, thin, align=center, fill=white},
  level-2/.style = {basic, rounded corners=5pt, thin,align=center, fill=white, text width=2.3cm},
  level-3/.style = {basic, thin, align=center, fill=white, text width=1.8cm},
  level-3-focus/.style = {basic, thick, align=center, fill=white, text width=1.8cm}
}
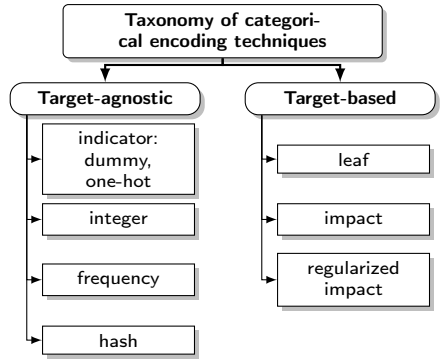
\begin{figure}[ht]
  \centering
  \begin{tikzpicture}[
    level 1/.style={sibling distance=8em, level distance=3em},
    edge from parent/.style={->,solid,black,thick,draw},
    edge from parent path={(\tikzparentnode.south) -- (\tikzchildnode.north)},
    >=latex, node distance=0.8cm, edge from parent fork down]
    \node[root] {\textbf{Taxonomy of categorical encoding techniques}}
      child {node[level-2, xshift=-10pt] (c1) {\textbf{Target-agnostic}}}
      child {node[level-2, xshift=10pt] (c2) {\textbf{Target-based}}};

    \begin{scope}[every node/.style={level-3}]
      \node [below of = c1, xshift=5pt] (c11) {indicator: dummy, one-hot};
      \node [below of = c11] (c12) {integer};
      \node [below of = c12] (c13) {frequency};
      \node [below of = c13] (c14) {hash};

      \node [below of = c2, xshift=5pt](c21) {leaf};
      \node [below of = c21] (c22) {impact};
      \node [below of = c22] (c23) {regularized impact};
    \end{scope}

    \foreach \value in {1,2,3,4}
      \draw[->] (c1.192) |- (c1\value.west);

    \foreach \value in {1,...,3}
      \draw[->] (c2.192) |- (c2\value.west);
  \end{tikzpicture}
  \caption{Taxonomy of common categorical variable encoding techniques.}
  \label{fig:overview}
\end{figure}

\emph{Target-based} methods try to incorporate information about the target values associated with a given level.
Early strategies aimed to reduce the number of levels by methods like hierarchical clustering or decision trees based on statistics of the target variable, although this has been rarely described in the scientific literature (for a brief mention, see Micci-Barreca 2001).
The basic idea of more advanced methods called target, impact, mean, or likelihood encoding is to use the training set to make a simple prediction of the target for each level of the categorical feature, and to use the prediction as the numerical feature value \(\hat{x}_l\) for the respective level.
An early formal description of this strategy is Micci-Barreca (2001).
In simple target encoding for regression problems, the mean target value in the training set from all observations with a certain feature level is used to encode that level for all observations: \(\hat{x}_l = \frac{\sum_{i:x^{train}_i = l}y^{train}_i}{N_l}\).
Simple target encoding often does not perform well with rare levels, where it tends to overfit to the training data and fails to generalize well for new observations.
In the extreme case of a categorical feature with unique values (e.g.~some hashed ID variable) studied in Prokhorenkova et al. (2018), the mean target for each level of this feature in simple target encoding is similar to the true target value of a single observation.
Based on the encoded feature, all observations can be predicted perfectly in the training set, even if the original variable did not contain any useful information.
ML models would place a high priority on such an encoded feature during training but would perform badly on test data.
To avoid this, practitioners often use regularized target encoding with a smoothing parameter that shrinks those effects towards the global mean (Micci-Barreca 2001).
An alternative strategy is to combine target encoding with cross-validation (CV) techniques (Prokhorenkova et al. 2018).

\hypertarget{categorical-encoding-benchmarks}{%
\subsection{Categorical Encoding Benchmarks}\label{categorical-encoding-benchmarks}}

Several small-scale studies have been previously conducted.
Those works did not yield conclusive results due to narrower scopes or not considering high cardinality variables.
One benchmark (\(6\) datasets) on encoding high cardinality features (cardinality between \(103\) and \(9095\)) in combination with gradient boosting has been published on the Kaggle forums (Prokopev 2018).
Different versions of target encoding are compared with indicator, integer, and frequency encoding.
They recommend combining the smoothed version of target encoding with 4- or 5-fold CV, never using simple target encoding and using indicator encoding only for small datasets.
Interestingly, frequency encoding did perform well in many cases.
Coors (2018) performed a benchmark (\(12\) datasets) on encoding high cardinality features (maximum number of levels per dataset between 10 and 25847) when developing the automatic gradient boosting (autoxgboost) library (Thomas, Coors, and Bischl 2018). They compared different variants of target encoding with integer and indicator encoding. In their benchmarks, target encoding only improved over target-agnostic methods on \(2\) datasets, while it led to worse results on 4 datasets. For a smaller number of levels, indicator and integer encoding yielded similar results.
As those studies only consider gradient boosting and a limited amount of datasets, it is unclear whether results generalize to other datasets and ML algorithms.
A recent study (\(15\) datasets) found good performance of target encoding, but they only investigated categorical features in regression settings (Seca and Mendes-Moreira 2021).
Several other publications studied encoding text data based on similarity (Patricio Cerda, Varoquaux, and Kégl 2018; P. Cerda and Varoquaux 2020) and employed indicator or target encoding as baselines.
Another line of work studies variable encodings employed within specific ML models.
Wright and König (2019) study treatments for categorical variables in random forests together with dummy and integer encoding (18 datasets, cardinality between \(3\) and \(38\)), concluding that indicator and integer encoding perform subpar in comparison to methods that re-order levels according to the target variable.
Prokhorenkova et al. (2018) compared their new CatBoost variant of target encoding to smoothed target encoding without CV, hold-out, and leave-one-out CV on \(8\) datasets.
While the CatBoost method performed best, hold-out came second and target encoding without CV performed worst.
An overview of encoding techniques tailored towards neural networks (e.g.~the widely adopted \emph{entity embeddings} by Guo and Berkhahn 2016) is provided in Hancock and Khoshgoftaar (2020).
They present a survey of indicator- and embedding-based methods but no benchmark study.
In contrast, our work is the first to focus on high cardinality variables and techniques that are agnostic to the subsequent ML method.
We study this problem on a larger variety of datasets and settings.
\newline

The main goal of our study is to assess the impact different categorical encoding techniques have on subsequent models' predictive performance.
As the optimal encoding might differ depending on the ML algorithm, we consider various state-of-the-art algorithms, \emph{regularized linear models} (LASSO), \emph{random forests} (RF), \emph{gradient tree boosting} (GB), \emph{k-nearest neighbors} (KNN), and \emph{support vector machines} (SVM).
To find default settings for high cardinality features, we analyze a variety of datasets with different characteristics including regression, binary classification, and multiclass classification problems.
Because our methods vary in runtime and complexity, we are interested in whether more complex methods are to be preferred, or if simpler approaches suffice.
We also study the relationship between a feature's cardinality and the choice of encoding technique by varying the minimum number of levels above which features are transformed.

\hypertarget{contributions}{%
\subsection{Contributions}\label{contributions}}

We survey a broad set of categorical encoding techniques and conduct a comprehensive benchmark study with a focus on high cardinality features.
We carefully design a benchmark scenario as well as a preprocessing scheme allowing us to study \(7\) different encoding techniques in conjunction with \(5\) commonly used ML algorithms across \(24\) diverse datasets, both from a classification and a regression regime.
We give a detailed description of our study design to highlight important considerations for studying high cardinality features.
Our results provide an overview of the performance of various approaches heavily used in the literature.
After a discussion of results concerning predictive performance, we provide further analyses seeking to inform practitioners which methods to apply.
This includes an important discussion of runtimes.

\hypertarget{encodings}{%
\section{Encodings}\label{encodings}}

\label{sec:methods}

Pseudocode for all encoders is presented in the Supplementary Material.
An important detail is how encoding techniques treat new levels during the prediction phase.

\hypertarget{integer-encoding}{%
\subsection{Integer Encoding}\label{integer-encoding}}

The simplest strategy for categorical features is integer encoding (also called ordinal encoding).
Observed levels from the training set are mapped to the integers \(1\) to \(L\).
Although new levels could be mapped to \(L+1\) or \(0\), model predictions would be arbitrary as the integer order does not carry information.
Thus, we encode new levels as missing values and use mode imputation to obtain the integer which matches the most frequent level in the training set.
Integer encoding should only be an acceptable strategy for tree-based models, which can separate all original levels with repeated splits.

\hypertarget{frequency-encoding}{%
\subsection{Frequency Encoding}\label{frequency-encoding}}

Frequency encoding maps each level to its observed frequency in the training set (\(\hat{x}_l = N_l\)).
This assumes a functional relationship between the frequency of a level and the target.
It implicitly reduces the number of levels, and the subsequent model can best differentiate between levels with dissimilar frequencies.
This approach is heavily used in natural language processing to encode token or n-gram counts.
We encode new levels with a frequency of \(1\).

\hypertarget{indicator-encoding}{%
\subsection{Indicator Encoding}\label{indicator-encoding}}

We use indicator encoding as an umbrella term for two common strategies to encode categorical features with a small to moderate number of levels: one-hot and dummy encoding.
One-hot encoding transforms the original feature into \(L\) binary indicator columns, each representing one original level.
An observation is coded with \(1\) for the indicator column representing its level (\(x^{train}_i = l\)) and \(0\) for all other indicators.
Dummy encoding results in only \(L-1\) indicator columns.
A reference feature level is chosen that is encoded with \(0\) in all indicator columns.
For one-hot encoding, the zero vector can be used to encode new levels which were not observed during training.
For dummy encoding, it is not useful to collapse new levels to the often arbitrary reference category; in our case the first level in alphabetical order.
We replace new levels in the prediction phase with the most frequent level in the training set.
As constructing all indicator variables is practically infeasible for high cardinality variables, we limit their number by collapsing rare levels beyond a varying threshold to a single \emph{other} category before encoding.

\hypertarget{hash-encoding}{%
\subsection{Hash Encoding}\label{hash-encoding}}

Hash Encoding can be used to compute indicator variables based on a hash function (Weinberger et al. 2009).
The basic idea is to transform each feature level \(l\) into an integer \(hash(l) \in \mathbb{N}\), based on its label.
This integer is then transformed into an indicator representation, with \(1\) in indicator column number \((hash(l) \mod hash.size) + 1\) and \(0\) in all remaining columns (Kuhn and Johnson 2019).
Some levels are hashed to the same indicator representation.
The smaller the \(hash.size\), the higher the number of collapsed levels.
The number of indicators is often effectively lower than \(hash.size\), as some indicators can be constant in the training set (we remove those columns for both training and prediction).
Although we could jointly hash multiple features, we hash each feature separately to improve comparability with the other encoders.

\hypertarget{leaf-encoding}{%
\subsection{Leaf Encoding}\label{leaf-encoding}}

Leaf encoding fits a decision tree on the training set to predict the target based on the categorical feature.
Each level is encoded by the number of the terminal node, in which an observation with the respective level ends up.
In that way, leaf encoding combines feature levels with similar target values.
We use the rpart package in R (Therneau and Atkinson 2018) which grows CARTs with categorical feature support that can be pruned based on internal performance estimates from 10-fold CV.
Thus, our leaf encoder automatically uses an ``optimal'' number of new levels.
To speed up the computation for multiclass classification, our implementation uses the ordering approach presented in Wright and König (2019).
New levels are encoded with the arbitrary number of the terminal node with most observations during training.
Encoded values are treated as a new categorical feature and encoded by one-hot encoding.
Our leaf encoder can be thought of as a simplification of the approach suggested by Grąbczewski and Jankowski (2003).

\hypertarget{impact-encoding}{%
\subsection{Impact Encoding}\label{impact-encoding}}

An early formal description of a so-called impact, target or James-Stein encoder was provided by Micci-Barreca (2001).
The basic idea is to encode each feature level with the conditional target mean (regression) or the conditional relative frequency of one or more target classes (classification).
The impact encoder for classification uses a logit link and transforms the original feature into \(C\) numeric features, each representing one target class.
A smoothing parameter \(\epsilon\) is introduced to avoid division by zero.
This parameter could be used to further regularize towards the unconditional mean.
We choose a small \(\epsilon = 0.0001\) as we want to compare simple target encoding with the regularized encoder introduced next.
Weights of evidence encoding from the credit scoring classification literature (Hand and Henley 1997) is almost identical to impact encoding, but without regularization or centering.

\hypertarget{glmm-encoding}{%
\subsection{GLMM Encoding}\label{glmm-encoding}}

Smoothed target encoding (Micci-Barreca 2001) can be interpreted as a simple (generalized) linear mixed model (glmm) in which the target is predicted by a random intercept for each feature level in addition to a fixed global intercept.
This connection is described in Kuhn and Johnson (2019).
To achieve regularized impact encoding, we implemented glmm encoders for regression, binary, and multiclass classification.
The encoded value for each level is based on the spherical conditional mode estimates.
In regression, the conditional modes are similar to the mean target value for each level, weighted by the relative observed frequency of that level in the training set (Gelman and Hill 2006; Bates 2020).
The estimate of the fixed intercept can be used during the prediction phase to encode new feature levels not observed in the training set.
In multiclass classification, we fit \(C\) one vs.~rest glmms resulting in one encoded feature per class.
An important advantage of using a glmm over impact encoding with a smoothing parameter is that a reasonable amount of regularization is determined automatically and tuning the complete ML pipeline is not necessary.

Overfitting can be further avoided through combination with cross-validation (CV) to train the encoder on independent observations without limiting the data to train the ML model.
We provide an implementation which combines target encoding based on glmms with CV.
During the training phase, we partition the data using CV into \(n.folds\) and fit a glmm on each resulting training set.
For each observation, there is exactly one glmm that did not use that observation for model fitting and can be safely used for encoding.
Note that the \(n.folds\) CV models (for \(n.folds\) \textgreater{} 1) are only used during the training phase.
In the prediction phase, feature values are always encoded by a single glmm fitted to the complete training set.
We study this method in three different settings: without CV (noCV), with \(5-\) (5CV) and with \(10-\) fold CV (10CV).
In our study, we use the lmer (regression) and glmer (classification) functions from the lme4 package in R (Bates et al. 2015) as an efficient way to fit glmms.

\hypertarget{control-conditions}{%
\subsection{Control Conditions}\label{control-conditions}}

We include three control conditions to better understand the effectiveness of the investigated encoders:
The performance of a featureless learner (\textbf{FL} condition) was estimated as a conservative baseline for each dataset.
In regression problems, FL predicts the mean of the target variable in the training set for each observation in the test set.
In classification problems, the most frequent class of the target within the training set is predicted.
For each dataset, we also consider a RF without encoding (\textbf{none} condition), to compare the use of encoding methods with a natural categorical splitting approach.
The ranger (Wright and Ziegler 2017) implementation provides efficient categorical feature support by ordering levels once before starting the tree growing algorithm (Wright and König 2019).
In the \textbf{remove} high cardinality features control condition, we omit features with a high number of levels above some threshold and use one-hot encoding (without collapsing rare levels) for the remaining features.
This condition reflects on whether including high cardinality features does indeed improve predictive performance.
Otherwise, the best encoding might just provide the least impairment compared to not including any high cardinality features.
We include an overview of available implementations in widely used ML frameworks for R and python in the Supplementary Material.

\hypertarget{benchmark-setup}{%
\section{Benchmark Setup}\label{benchmark-setup}}

\hypertarget{datasets}{%
\subsection{Datasets}\label{datasets}}

A table showing a detailed summary of all benchmark datasets can be found in the Supplementary Material.
We specifically investigate datasets that contain categorical variables with a large number of levels, including many well-known datasets
from previous studies (Patricio Cerda, Varoquaux, and Kégl 2018; Prokhorenkova et al. 2018; Coors 2018; Kuhn and Johnson 2019).
All datasets can be downloaded from the \texttt{OpenML} platform (Vanschoren et al. 2013) based on the name or the displayed OmlId.
The datasets include 8 regression, 10 binary classification, and 6 multiclass classification problems (between 3 and 12 classes).
To assess the imbalance in categorical variables we computed the normalized entropy for each categorical variable.
The maximum normalized entropy is \(1\), which corresponds to a uniform distribution, while a lower number indicates a larger imbalance.
Sample sizes range between 736 and 1224158 observations.
The total number of features ranges between 5 and 208.
Datasets contain between 1 and 20 categorical features with more than \(10\) levels, with the maximum number of levels for a feature ranging between 14 and 30114. Missing values are present in about half of the datasets.

\hypertarget{high-cardinality-threshold}{%
\subsection{High Cardinality Threshold}\label{high-cardinality-threshold}}

\label{sec:HCT}

It is often assumed that advanced encoding methods are only advantageous for variables with a high number of levels, while simple indicator encoding is more appropriate for a small number of levels.
To reflect this in our benchmark, a high cardinality threshold (HCT) parameter with values of \(10\), \(25\), and \(125\) was introduced determining varying configurations for the different encoders.
For indicator encoding, the \(HCT - 1\) most frequent levels are encoded together with a single collapsed category for the remaining levels.
For integer, frequency, hash, leaf, impact, and glmm encoders, only features with more than HCT levels in the training set were encoded with the respective strategy, while the remaining categorical features were one-hot encoded. HCT is used as the hash size in hash encoding.
In the remove control condition, features with more than HCT levels in the training set are removed from the feature set.\footnote{Based on the number of categorical features and levels per feature, some HCT settings were removed from the benchmark for some combinations of dataset $\times$ encoder.
This ensured that encoders always affect at least one feature and that the remove condition always removes at least one feature.
If settings where an encoder would lead to identical encoding strategies for all features of a dataset, we only kept the condition with the smallest HCT value.}

\hypertarget{machine-learning-pipeline-and-algorithms}{%
\subsection{Machine Learning Pipeline and Algorithms}\label{machine-learning-pipeline-and-algorithms}}

\label{sec:pipeline}

In total, we investigate \(5\) ML algorithms. We kept tuning their hyperparameters to a minimum because we were interested in the effect of the encoding techniques instead of a comparison of ML algorithms.
LASSOs were fitted with \texttt{glmnet} (Friedman, Hastie, and Tibshirani 2010), internally tuning the regularization using 5-fold CV.
RFs with \(500\) trees were trained using \texttt{ranger} (Wright and Ziegler 2017) without tuning, because RFs can be expected to give reasonable results with default settings (Probst, Wright, and Boulesteix 2019).
GB models were trained using \texttt{xgboost} (Chen et al. 2018), setting the learning rate to \(0.01\) and determining the number of iterations using early stopping on a 20\% holdout set.
KNN was taken from package \texttt{kknn} (Schliep and Hechenbichler 2016) standardizing features and using a constant \(k=15\) for the number of nearest neighbors, together with an information gain filter (Brown et al. 2012) to limit the number of features to \(25\).
SVMs with radial basis function kernel were trained with \texttt{liquidSVM} (Steinwart and Thomann 2017).
The bandwidth and regularization parameters were internally tuned using 5-fold CV. We used a one-vs-all approach for multiclass-settings.

The ML pipeline outlined below was used for all experimental conditions.
It was carefully designed to ensure consistent results for extreme conditions (e.g.~some levels only existing in the training data).

\textbf{Imputation I}: Create a new factor level for missing values in categorical features with more than two categories. Impute missing values in binary features using the mode and missing values in numerical features using the mean feature value in the training data.

\textbf{Encoding}: Transform the complete categorical features by the respective encoder.
In the \textit{no encoding} condition, the encoder simply passes on its input.
The leaf and remove conditions still return categorical features, while the remaining encoders return only numerical features. Encoders only affect categorical variables above the specified HCT value.

\textbf{Imputation II}: To handle new levels observed during prediction, impute missing values obtained during encoding.

\textbf{Drop constants}: Drop features that are constant during training. As none of the original datasets includes constant columns, this step only removes constant features that are produced by the encoders or the CV splitting procedure.

\textbf{Final one-hot encoding}: Transform all remaining
categorical features via one-hot encoding (skipped for no encoding condition).

\textbf{Learner}: Use the transformed data from each training set to fit the respective ML algorithm.
In the prediction phase, transformed feature values (based on the trained encoder) for new observations in each test set are fed into the trained model to compute predictions.

\hypertarget{performance-evaluation}{%
\subsection{Performance Evaluation}\label{performance-evaluation}}

We perform all analyses in the open-source statistical software R (R Core Team 2021).
To enable a fair and reliable comparison in our study, we implemented all encoding methods on top of the \texttt{mlrCPO} package (Binder 2018).
The pre-processing, as well as the final ML algorithm described in \ref{sec:pipeline}, were trained and resampled using the \texttt{mlr} framework (Bischl et al. 2016) together with the \texttt{batchtools} package (Lang, Bischl, and Surmann 2017) to scale the benchmark analysis to HPC compute infrastructure.
All materials for this study, including reproducible code for this manuscript and result objects can be downloaded from our \textbf{online repository} \footnote{\url{https://github.com/slds-lmu/paper_2021_categorical_feature_encodings}}.

Throughout our experiments, we use 5-fold CV to obtain estimates of predictive performance.
Depending on the target variable, we report root mean squared error (RMSE) for regression, area under the curve (AUC) for binary classification, and its extension AUNU (Ferri, Hernández-Orallo, and Modroiu 2009) for multiclass problems.

In our benchmark study, each encoding method listed in Section \ref{sec:methods} is combined with all ML algorithms (c.f. Section \ref{sec:pipeline}) across
high cardinality thresholds (HCT) \(10\), \(25\), \(125\).
While we only report results for the best HCT setting for each ML algorithm \(\times\) dataset combination in Section \ref{sec:results}, we study the effect of the HCT parameter in more detail in section \ref{sec:hct_study}.

\hypertarget{benchmark-results}{%
\section{Benchmark Results}\label{benchmark-results}}

\label{sec:results}

Our main question is which encoding methods generally work well across various datasets.
We report results for regression and classification datasets separately, as the associated metrics differ in scale.

For 6 datasets, some conditions with the SVM led to unexpected crashes due to memory problems or numerical errors.
We completely removed those datasets for the SVM when computing ranks or other statistics that compare encodings across datasets.

\hypertarget{encoder-performance}{%
\subsection{Encoder Performance}\label{encoder-performance}}

\begin{figure}[!h]
\includegraphics[width=\textwidth]{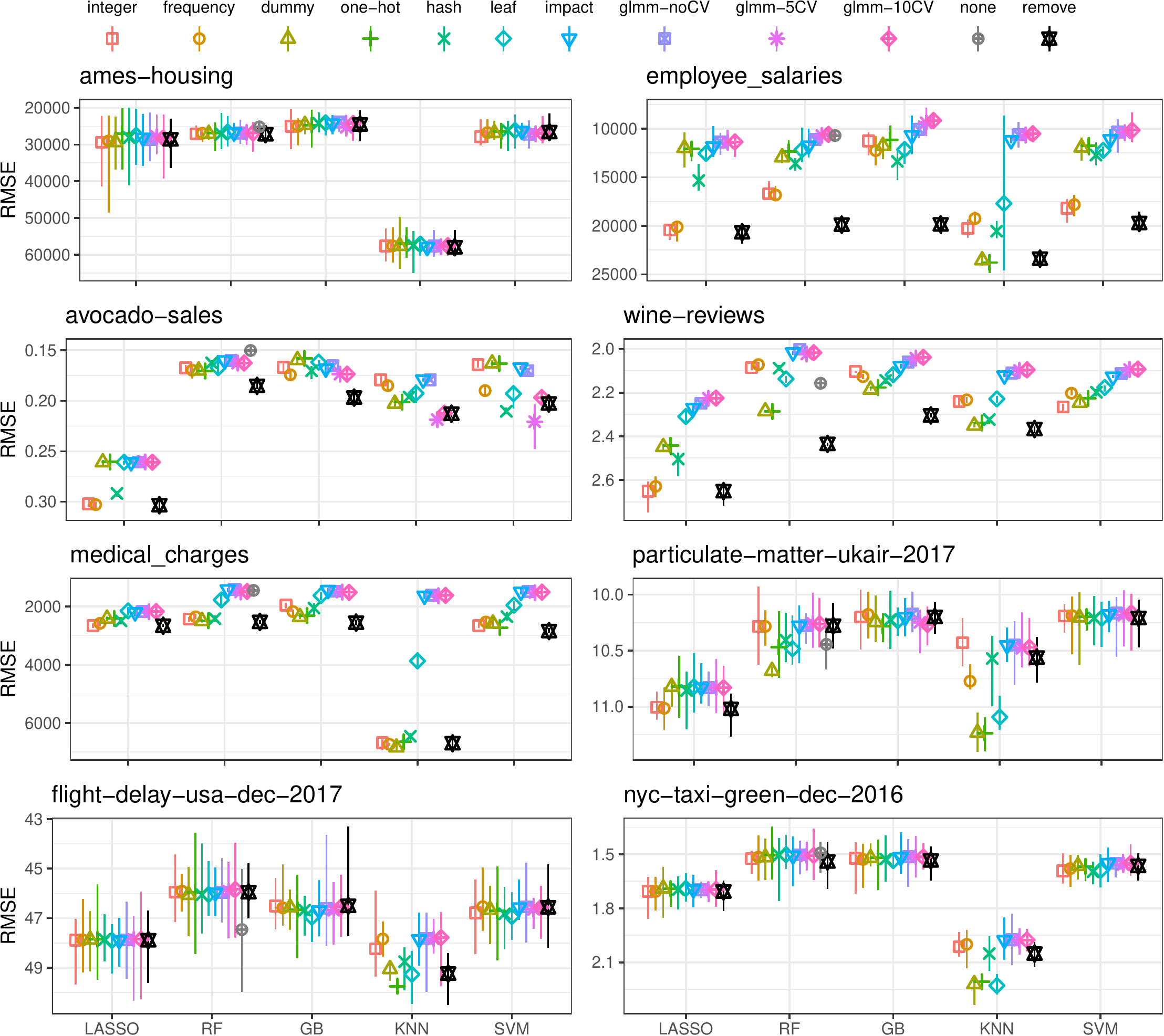} \caption{Performance estimates from 5-CV for regression (mean, min, max). For each combination, only the best HCT condition is displayed. Note the reversed y-axis to ease visual interpretation.}\label{fig:regr-res}
\end{figure}

\begin{figure}[!h]
\includegraphics[width=\textwidth]{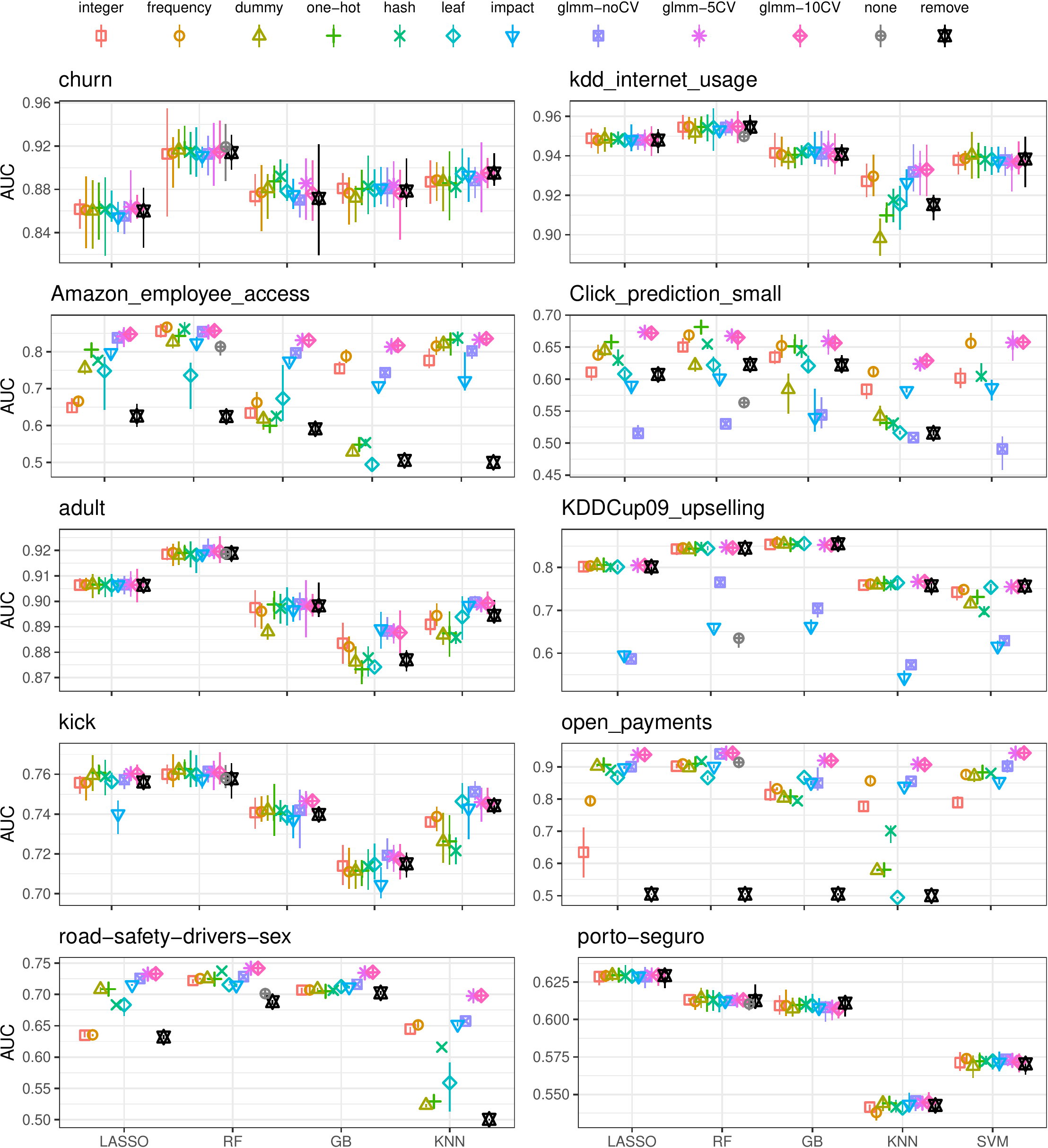} \caption{Performance estimates from 5-CV for binary classification (mean, min and max). For each combination, only the best HCT condition is displayed.}\label{fig:bincl-res}
\end{figure}

\begin{figure}[!h]
\includegraphics[width=\textwidth]{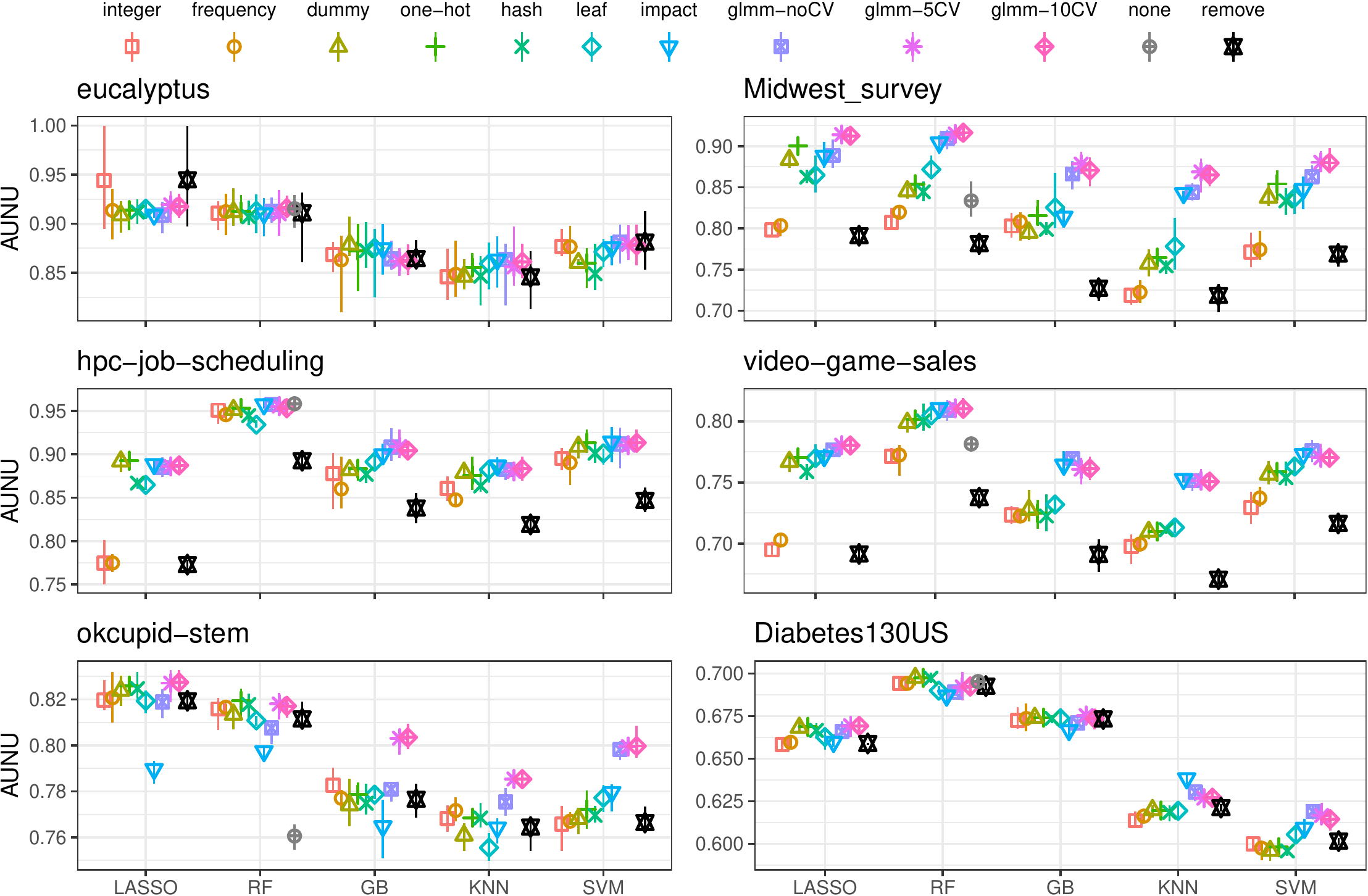} \caption{Performance estimates from 5-CV for multiclass classification (mean, min and max). For each combination, only the best HCT condition is displayed.}\label{fig:multcl-res}
\end{figure}

Mean performance estimates along with minimum and maximum performance are reported for all datasets in Figures \ref{fig:regr-res} (regression), \ref{fig:bincl-res} (binary classification), and \ref{fig:multcl-res} (multiclass classification).
To reduce the complexity induced by the hyperparameter HCT we only display the parameter condition with the best performance for each combination of dataset \(\times\) encoding \(\times\) ML algorithm.
The y-axis differs for all datasets and is reversed for the RMSE for better visual comparison.
For some datasets, the remove condition performed very similar to the other encodings (e.g.~\emph{ames-housing}, \emph{porto-seguro}), suggesting that categorical features were not informative.
Performance in some CV folds was below the FL learner for \emph{flight-delay-usa-dec-2017} (\(RMSE_{FL} =\) 48.81) and \emph{nyc-taxi-green-dec-2016} (\(RMSE_{FL} =\) 2.22).

On datasets with substantive performance differences, target encoding with the glmm encoder was generally \textbf{most} effective.
The \textbf{worst} encoder differed by ML algorithm and dataset.
For datasets \emph{Click\_prediction\_small}, \emph{KDDCup09\_upselling}, \emph{kick}, and \emph{okcupid-stem} some encoder \(\times\) ML algorithm conditions performed worse than simply removing high cardinality features.

\hypertarget{meta-rankings-and-dataset-clustering}{%
\subsection{Meta Rankings and Dataset Clustering}\label{meta-rankings-and-dataset-clustering}}

\begin{figure}
\includegraphics[width=\textwidth,height=.55\textwidth]{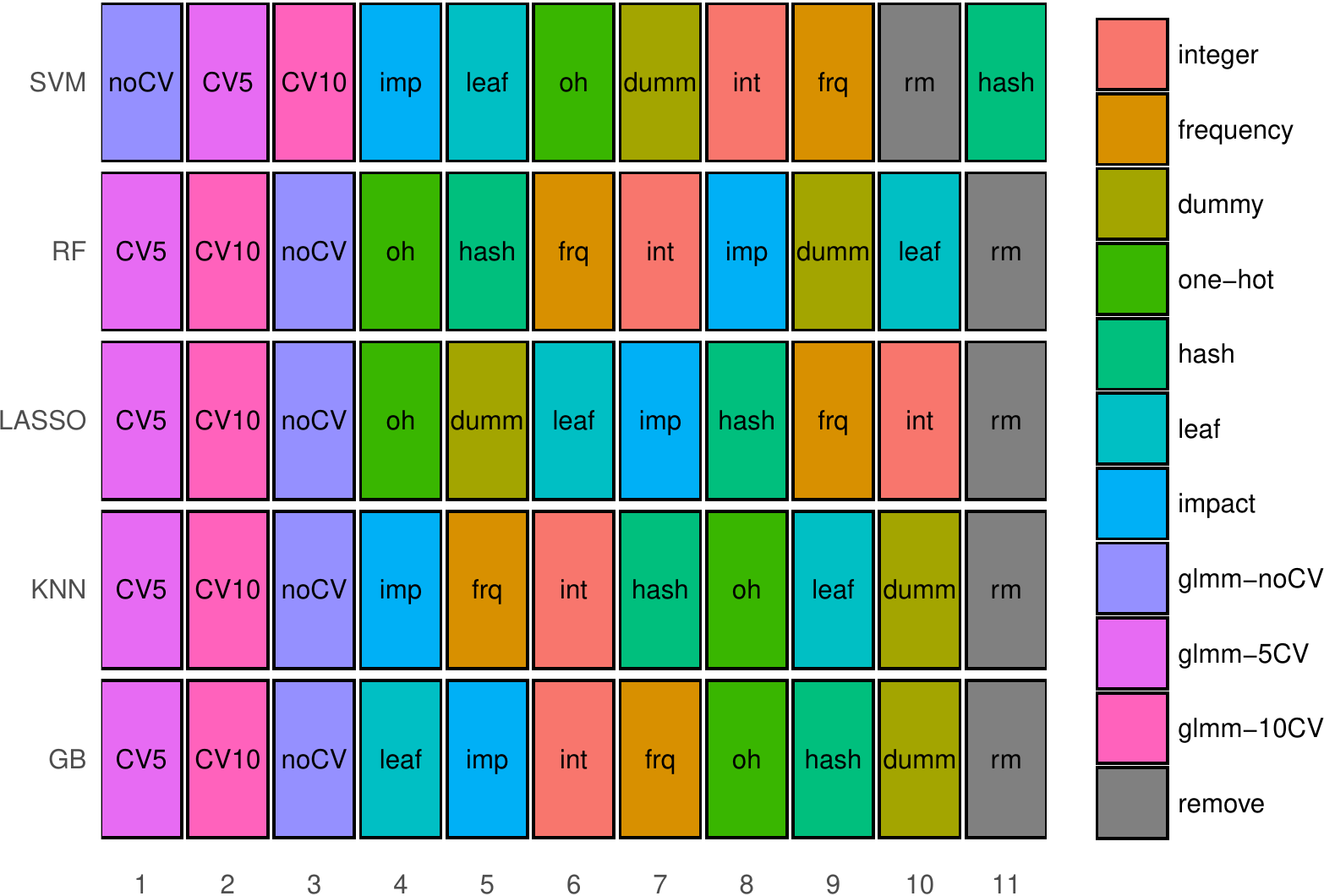} \caption{Consensus rankings across all datasets for each algorithm. Lower ranks indicate better performance. The rank of the \emph{none} control condition of the RF (rank 11 of 12) was omitted from the figure.}\label{fig:rank-res}
\end{figure}

To further analyze our results, we used statistical inference methods inspired by the benchmark community in computational statistics (e.g., Hothorn et al. 2005).
First, we present meta rankings for each ML algorithm in Figure \ref{fig:rank-res}:
We defined an encoder relation within each dataset, based on corrected resample t-tests (Nadeau and Bengio 2003).
An encoding was defined to beat another encoding if the one-sided p-value of the t-test was \(< .05\).
This allowed us to compute a weak-order consensus ranking \(R\) defined by the optimization problem:

\[ \underset{R \in \mathcal{C}}{arg\ min} \ \sum_{b=1}^B d(R_b, R)\]

where \(d\) is the symmetric difference distance and \(R_b\) is the relation for dataset \(b\) (Hornik and Meyer 2007; Meyer and Hornik 2018).
The symmetric difference between two relations is the number of cases one encoding beats another encoding in one relation but not in the other one.

Although the presented solutions of the optimization problem are not unique, rankings were highly stable for the high and low ranks.
Meta rankings seem to be highly consistent with the individual patterns of encoder performances reflected in Figures \ref{fig:regr-res} to \ref{fig:multcl-res}.
Looking at meta-rankings, approaches based on GLMM's in combination with cross-validation outperform all other approaches across all algorithms.
A further interesting detail omitted in Figure \ref{fig:rank-res} for clarity is that the \textit{none} encoding strategy for the RF was beaten by all strategies except for the \textit{remove} condition.
This implies, that even if the algorithm provides a mechanic for treating categorical variables, it might often be optimal to use a different strategy instead.

\begin{figure}[!h]
\includegraphics[width=\textwidth]{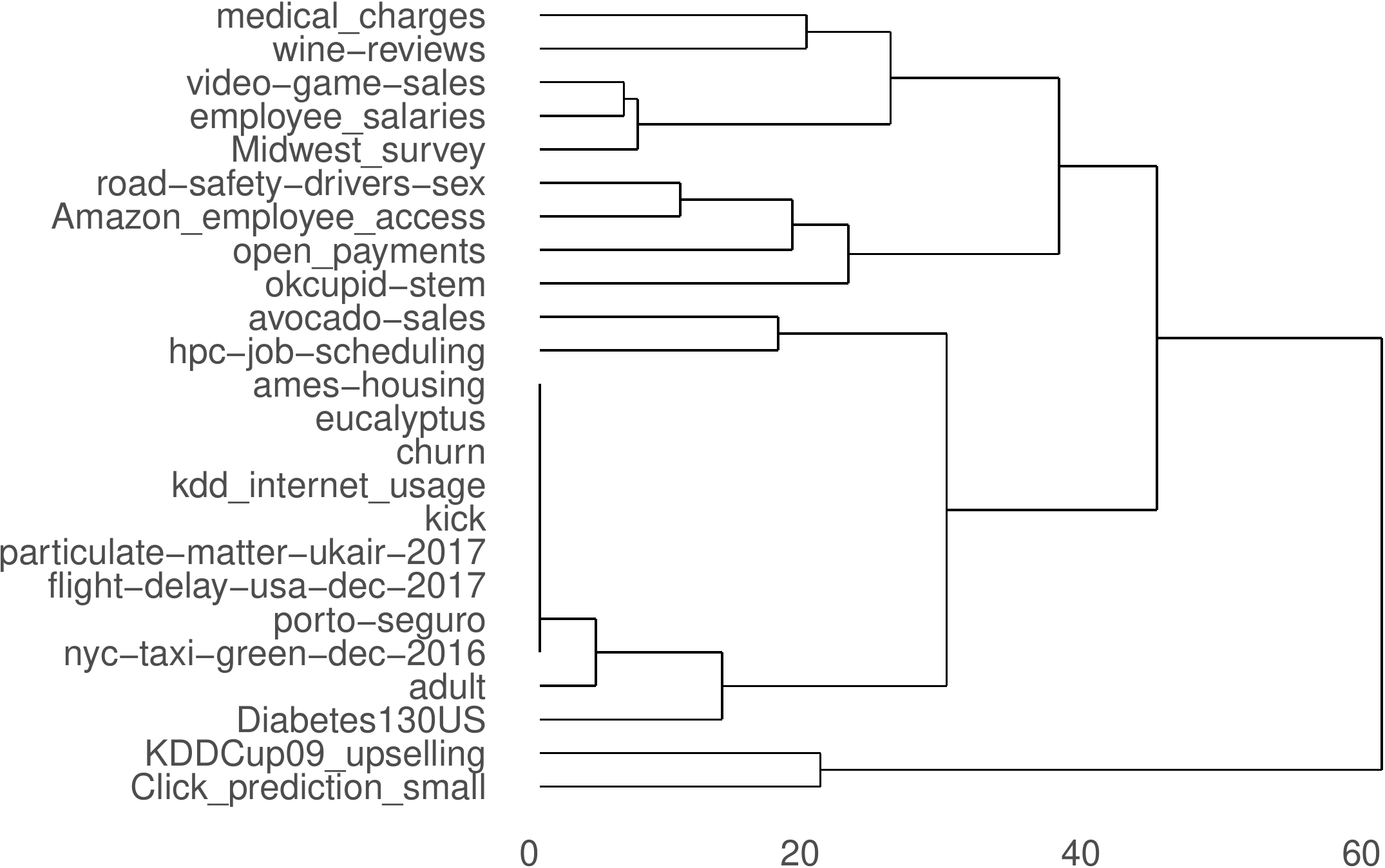} \caption{Hierarchical cluster analysis of benchmark datasets. The symmetric difference distance between two datasets reflects differences in performance patterns between encodings (independent of the employed ML algorithm).}\label{fig:dendrogram}
\end{figure}

In a second exploratory analysis, we tried to find clusters of datasets based on systematic patterns of encoder performance (independent of the employed ML algorithm).
We computed a partial-order consensus relation across ML algorithms for each dataset and hierarchically clustered the dataset consensus relations using the symmetric difference distance in combination with the complete linkage agglomeration method.
The resulting dendrogram is displayed in Figure \ref{fig:dendrogram}.
Although the cluster structure is somewhat ambiguous, roughly three clusters can be described:
The first 9 datasets from the top of the dendrogram are characterized by a medium to high number of levels, low performance of the remove condition (indicating the importance of high cardinality features) and clear performance advantage of target encoders.
For the next 13 datasets which contain the smallest number of levels, traditional encodings can compete with target encoding.
This biggest cluster also includes 9 datasets with zero distances, in which no significant performance differences could be observed between any encoding conditions (nor with the remove condition, indicating that high cardinality features are less informative).
The last two datasets with the highest number of levels formed a separate cluster, in which target encoding without strong regularization (impact, glmm-noCV) showed severe overfitting.
Note that clusters were not determined by problem type, again suggesting that encoder rankings are somewhat similar for regression and classification settings.\footnote{We tried to corroborate this observation by comparing meta-rankings between problem types. Unfortunately, problem type specific consensus relations did not converge (probably due to the reduced number of datasets) and were uninterpretable.}

\hypertarget{summary-of-encoder-performance}{%
\subsection{Summary of Encoder Performance}\label{summary-of-encoder-performance}}

Regularized target encoding was superior or at least competitive on all datasets.
We could not observe a setting in which regularized target encoding was convincingly beaten by target agnostic methods.
Especially effective was glmm encoding with 5-fold-CV, which ranked first place for all ML algorithms except SVM.
Performance often did not improve with glmm-10CV, suggesting that 5 folds might be a good regularization default in practice.
In line with earlier research (Micci-Barreca 2001; Prokhorenkova et al. 2018), target encoding with regularization (glmm) performed better or equally well in comparison with the unregularized impact encoder.
When glmm performed better, impact encoding not only performed worse than glmm with CV but was sometimes also inferior compared to other encoders.

The following observations were also interesting:
Surprisingly, integer encoding did not perform well with GB and target-based encoders (especially the glmm encoder) seem to be preferable.
For LASSO, previous studies have suggested that indicator encoding works well, even with a very high number of levels (Patricio Cerda, Varoquaux, and Kégl 2018).
Although the glmm encoders ranked first in our benchmark for the LASSO, it was the only algorithm where the indicator encoders achieved the next best ranking.
Note that for computational reasons we limited the maximum amount of indicator variables per feature to 125 in our experimental design.
The HCT = 125 setting performed best for LASSO with indicator encoding in a large number of datasets, indicating that performance might have further improved with higher values.
Both KNN and SVM rely on numerical distances in feature space and tend to perform poorly in the presence of high dimensionality.
Thus, we expected that target encoding should work well here as it transforms categories into a single, smooth numerical feature.
This was backed by our benchmark results.
For some datasets (\emph{medical\_charges}, \emph{road-safety-drivers-sex}, and \emph{Midwest\_survey}) KNN (without tuning of the optimal number of nearest neighbors) could compete with the more sophisticated ML algorithms when combined with target encoding, but performed poorly with other encoders.
Although consistent with the big picture, SVM results have to be considered with care, as some experimental conditions resulted in unexpected computational errors.
In RF, a widely used strategy to deal with categorical features is to order levels by average target statistics for a given level.
For a small number of levels, this approach has been reported superior to indicator and integer encoding (Wright and König 2019), while we observed poor performance for datasets with a larger number of levels.

When looking for a simple default encoding, indicator encoding in combination with collapsing small levels seems a robust alternative, although the glmm encoders performed better.
We found that one-hot encoding usually gave a slightly better performance than dummy encoding, which has also been observed by Chiquet, Grandvalet, and Rigaill (2016) and Tutz and Gertheiss (2016) (p.254).
Our results suggest that one-hot encoding is the better standard compared to dummy encoding when applying nonlinear regularized models like RF, GB or SVM with (high cardinal) categorical features.
We provide a more detailed comparison for indicator encoding in the Supplementary Material.

\hypertarget{runtime-analysis}{%
\subsection{Runtime Analysis}\label{runtime-analysis}}

\begin{table}

\caption{\label{tab:time-ranks}Proportional Increase in Runtime Compared to One-Hot Encoding}
\centering
\begin{threeparttable}
\begin{tabular}[t]{l|l|l|l|l|l}
\hline
Encoding & LASSO & RF & GB & KNN & SVM\\
\hline
integer & $0.4_{0}^{1.4}$ & $0.59_{0}^{1.1}$ & $0.67_{0}^{0.9}$ & $0.85_{0.4}^{1.5}$ & $1.01_{0.6}^{1.6}$\\
\hline
frequency & $0.34_{0}^{2.1}$ & $0.54_{0}^{1.7}$ & $0.54_{0}^{1.5}$ & $0.79_{0.2}^{1.1}$ & $1.11_{0.5}^{1.7}$\\
\hline
dummy & $1.13_{0.8}^{2.8}$ & $0.91_{0.5}^{1.7}$ & $0.97_{0.4}^{5.7}$ & $1.05_{0.6}^{1.6}$ & $1.02_{0.7}^{1.5}$\\
\hline
one-hot & $1_{1}^{1}$ & $1_{1}^{1}$ & $1_{1}^{1}$ & $1_{1}^{1}$ & $1_{1}^{1}$\\
\hline
hash & $0.87_{0.2}^{2.6}$ & $1.03_{0.4}^{3.3}$ & $0.98_{0.4}^{7.3}$ & $0.93_{0.5}^{2.2}$ & $1.16_{0.5}^{2}$\\
\hline
leaf & $0.46_{0}^{1.7}$ & $0.67_{0}^{2.3}$ & $0.85_{0}^{9.1}$ & $0.97_{0.5}^{2.2}$ & $1.01_{0.6}^{1.9}$\\
\hline
impact & $0.5_{0.1}^{1.9}$ & $0.6_{0.1}^{1.5}$ & $0.82_{0}^{120.5}$ & $1.11_{0.3}^{24}$ & $1.06_{0.7}^{1.8}$\\
\hline
glmm-noCV & $0.58_{0}^{2.8}$ & $0.57_{0.1}^{3.8}$ & $1.65_{0}^{12}$ & $1.09_{0.3}^{10.3}$ & $1.1_{0.7}^{2.1}$\\
\hline
glmm-5CV & $0.83_{0.1}^{2.2}$ & $0.66_{0.1}^{15.3}$ & $7.21_{0}^{69.3}$ & $2.77_{0.7}^{50.1}$ & $1.41_{0.6}^{4.6}$\\
\hline
glmm-10CV & $1.07_{0.1}^{4.2}$ & $0.96_{0.1}^{28.4}$ & $12.71_{0}^{127.6}$ & $5.37_{0.8}^{97.9}$ & $1.55_{0.5}^{4.6}$\\
\hline
none &  & $0.2_{0}^{1.4}$ &  &  & \\
\hline
remove & $0.4_{0}^{1.7}$ & $0.48_{0}^{1.1}$ & $0.58_{0}^{1.3}$ & $0.82_{0.2}^{1.2}$ & $1.09_{0.6}^{2}$\\
\hline
\end{tabular}
\begin{tablenotes}
\item \textit{Note: } 
\item Median $_{min}^{max}$ across datasets of the proportional increase in runtime from 5-CV, when comparing the respective encoder with one-hot encoding. Only the best HCT conditions are reported.
\end{tablenotes}
\end{threeparttable}
\end{table}

To determine whether traditional encodings are preferred when facing limited computational resources, we further analyzed runtimes of the whole analysis pipeline for different encoders and ML algorithms.
Aggregated results are shown in Table \ref{tab:time-ranks}.
To enable a meaningful comparison, we report runtime as the fraction of a full pipeline's strategy compared to the one-hot encoding condition and then further aggregate across datasets using the median.
Again, we only report the \textbf{best} HCT setting.
Absolute runtimes are hard to interpret and aggregate because runtime distributions across datasets are heavily skewed as proportionally larger runtimes are observed for big datasets.
While a pipeline's runtime can be dominated by the encoder for small datasets, the training of the consecutive ML algorithm is dominating for large datasets, which might render differences during encoding irrelevant.
Therefore we aim to report what is important in practice, the time differences for training the \textbf{full pipeline}.
The results clearly show that regularized target encoding does not consistently yield slower runtimes compared to simple strategies like indicator encoding. Supposedly, the more efficient representations produced by target encoding lead to faster runtimes of subsequent ML algorithms.
This suggests that a possible runtime vs.~predictive performance trade-off might also be in favour of target encoding, especially for large datasets where a high number of indicator variables increases computational load.
We did observe a substantial increase in runtimes when using regularized glmm with GB (with little tuning), where runtimes are relatively short in comparison to the time required to fit categorical encoders.
In other settings (LASSO, RF), the glmm encoders have been observed to be even faster than indicator encoding.

\hypertarget{analysing-high-cardinality-thresholds}{%
\subsection{Analysing High Cardinality Thresholds}\label{analysing-high-cardinality-thresholds}}

\label{sec:hct_study}

Until now, we ignored the HCT parameter by reporting only the condition with the best performance.
An interesting question is whether target encoding is only useful for features with a very high number of levels or also for fewer levels, where most practitioners would routinely use indicator encoding.
We tested this using HCT thresholds of \(10\), \(25\) and \(125\), but the results are not easy to interpret.
In general, the optimal threshold seemed to strongly depend on the dataset, but we also observed some weak patterns:
LASSO improved for larger HCT, indicating that its internal regularization can efficiently deal with the sparseness induced by indicator encoding.
In comparison, other methods generally yielded better performance if features above the very low HCT of \(10\) were encoded using one of the target encoding strategies.
Differences in regularization for the target based encoders seemed not to prefer different HCT values.

\hypertarget{discussion}{%
\section{Discussion}\label{discussion}}

In our benchmark, we compared encoding strategies for high cardinality categorical features on a variety of regression, binary, and multiclass classification datasets with different ML algorithms.
Regularized target encoding was superior across most datasets and ML algorithms.
Although the performance of other encoding strategies was comparable in some conditions, target encoding was never outperformed.
In general, our results suggest that regularized target encoding based on glmms with 5-fold CV (glmm-5CV) works well for all kinds of algorithms and should be a reasonable default strategy.
It sometimes leads to slightly longer runtimes (in comparison to indicator encoding), but especially for larger datasets this is often offset by the more efficient representation produced by target encoding.
The glmm encoder has a clear advantage over target encoding with a smoothing hyperparameter (Micci-Barreca 2001), as costly tuning the whole ML pipeline with different smoothing values is not required.

What constitutes a ``high'' cardinality problem is a difficult question that might not only depend on the number of levels but also on other characteristics of the dataset and its features.
Supposedly, the number of datasets in our study was too small to discover consistent patterns between encoder performance and dataset characteristics.
Target encoding features with only 10 levels seemed to be effective in a substantive number of conditions, but for other datasets, higher HCT values performed better.
Thus, some form of hyperparameter tuning seems necessary to decide the level threshold for target encoding at this point, as no suitable defaults seem to be available.
Note that we compared different HCT values but then used the same encoding strategy for all affected features alongside one-hot encoding for the remaining ones.
A further improvement could be to decide whether to use target encoding on a feature by feature basis.
ML pipelines could introduce a categorical hyperparameter for each feature that represents which encoding is used.
To make this complicated meta optimization problem feasible, only a small number of encoders can be included.
Our study can help to decide which encoders could be safely omitted from consideration.

\hypertarget{limitations}{%
\subsection{Limitations}\label{limitations}}

Several design decisions were necessary to make answering our research questions computationally feasible:
We used minimal tuning for our ML algorithms, as we were not interested in comparing their performance against each other.
This probably led to suboptimal performance for the GB, KNN, and SVM learners.
When interpreting our results, we assume that the encoder rankings are comparable when more extensive tuning is used.
This is plausible because the meta rankings between algorithms were highly stable.
We only used 5-fold-CV without repetitions to estimate predictive performance, but due to a small variance between folds, we could confidently detect performance differences of encoders for many datasets.
Because we are interested in model-agnostic methods that can be combined with any supervised ML algorithm, we did not investigate recent model-specific strategies (Guo and Berkhahn 2016; Prokhorenkova et al. 2018).
We did not include deep neural networks in our benchmark because they are still rarely applied in the statistical learning community and more dominant in computer science.
We do not differentiate between nominal and ordinal categorical features:
Most publicly available datasets simply do not contain any high cardinal ordinal features.
For many applications, ordinal feature information is not available as metadata which makes it less relevant for some applications like automatic ML.
Thus, our benchmark does not include specific ordinal encoding methods like Helmert and Polynomial contrasts (Chambers and Hastie 1992) or ordinal penalization methods (Tutz and Gertheiss 2016).
We only investigate traditional categorical variables and do not extend our analysis to multi-categorical variables or text-strings (Patricio Cerda, Varoquaux, and Kégl 2018; P. Cerda and Varoquaux 2020), where other encoding techniques can harvest additional information.
We only investigate univariate encodings, i.e.~we always encode each variable separately.
Levels with comparable main effects can not be distinguished based on the transformed feature, which prevents the consecutive ML algorithm from learning interactions for specific levels.
Methods that jointly encode features and therefore leverage correlational structures between features are an interesting avenue for future research:
Recent approaches use target-based encoding strategies to learn a vector-valued representation of a level, similar to neural network embeddings (Guo and Berkhahn 2016; Rodríguez et al. 2018).
Those should be compared to classical approaches from the optimal scaling literature (e.g. De Leeuw, Young, and Takane 1976; Young, De Leeuw, and Takane 1976) or the subsequent \emph{aspect} framework (Mair and Leeuw 2010), which are unfortunately unfamiliar to many ML researchers.

\hypertarget{conclusion}{%
\subsection{Conclusion}\label{conclusion}}

This benchmark study compared the predictive performance of a variety of strategies to encode categorical features with a high number of unordered levels for different supervised ML algorithms.
Regularized versions of target encoding, which uses predictions of the target variable as numeric feature values, performed better than traditional strategies like integer or indicator encoding.
Most effective, with a consistently superior performance across ML algorithms and datasets, was a target encoder which combines simple generalized linear mixed models with cross-validation and does not require hyperparameter tuning.
GLMMs are a major workhorse in applied statistics but not well understood and often neglected by the ML community.
Refining target encoders that use statistical models for more efficient regularization and studying their theoretical properties could be a valuable research topic for the computational statistics community.

\hypertarget{declarations}{%
\subsection{Declarations}\label{declarations}}

\hypertarget{funding}{%
\subsubsection{Funding}\label{funding}}

This work has been funded by the German Federal Ministry of Education and Research (BMBF) under Grant No.~01IS18036A and by the Bavarian Ministry of Economic Affairs, Regional Development and Energy through the Center for Analytics -- Data -- Applications (ADA-Center) within the framework of „BAYERN DIGITAL II`` (20-3410-2-9-8).

\hypertarget{conflicts-of-interest}{%
\subsubsection{Conflicts of Interest}\label{conflicts-of-interest}}

The authors declare no conflict of interest.

\hypertarget{code-availability}{%
\subsubsection{Code Availability}\label{code-availability}}

All analysis code is publicly available at \url{https://github.com/slds-lmu/paper_2021_categorical_feature_encodings}.

\hypertarget{availability-of-data}{%
\subsubsection{Availability of Data}\label{availability-of-data}}

All benchmark datasets are publicly available at \url{https://www.openml.org/}.

\hypertarget{references}{%
\section*{References}\label{references}}
\addcontentsline{toc}{section}{References}

\raggedright

\hypertarget{refs}{}
\begin{CSLReferences}{1}{0}
\leavevmode\vadjust pre{\hypertarget{ref-bates_2018}{}}%
Bates, Douglas. 2020. {``Computational Methods for Mixed Models.''} \emph{Vignette for Lme4}. \url{https://cran.r-project.org/web/packages/lme4/vignettes/Theory.pdf}.

\leavevmode\vadjust pre{\hypertarget{ref-bates_2015}{}}%
Bates, Douglas, Martin Mächler, Ben Bolker, and Steve Walker. 2015. {``Fitting Linear Mixed-Effects Models Using {lme4}.''} \emph{Journal of Statistical Software} 67 (1): 1--48. \url{https://doi.org/10.18637/jss.v067.i01}.

\leavevmode\vadjust pre{\hypertarget{ref-binder_2018}{}}%
Binder, Martin. 2018. \emph{mlrCPO: Composable Preprocessing Operators and Pipelines for Machine Learning}. \url{https://github.com/mlr-org/mlrCPO}.

\leavevmode\vadjust pre{\hypertarget{ref-bischl_2016}{}}%
Bischl, Bernd, Michel Lang, Lars Kotthoff, Julia Schiffner, Jakob Richter, Erich Studerus, Giuseppe Casalicchio, and Zachary M. Jones. 2016. {``{mlr}: Machine Learning in r.''} \emph{Journal of Machine Learning Research} 17 (170): 1--5. \url{http://jmlr.org/papers/v17/15-066.html}.

\leavevmode\vadjust pre{\hypertarget{ref-bommert}{}}%
Bommert, Andrea, Xudong Sun, Bernd Bischl, Jörg Rahnenführer, and Michel Lang. 2020. {``Benchmark for Filter Methods for Feature Selection in High-Dimensional Classification Data.''} \emph{Computational Statistics \& Data Analysis} 143 (C). \url{https://doi.org/10.1016/j.csda.2019.106839}.

\leavevmode\vadjust pre{\hypertarget{ref-boulesteix_2017}{}}%
Boulesteix, Anne-Laure, Harald Binder, Michal Abrahamowicz, Willi Sauerbrei, et al. 2017. {``On the Necessity and Design of Studies Comparing Statistical Methods.''} \emph{Biometrical Journal. Biometrische Zeitschrift} 60 (1): 216--18. \url{https://doi.org/10.1002/bimj.201700129}.

\leavevmode\vadjust pre{\hypertarget{ref-brown2012}{}}%
Brown, Gavin, Adam Pocock, Zhao Ming-Jie, and Mikel Luján. 2012. {``Conditional Likelihood Maximisation: A Unifying Framework for Information Theoretic Feature Selection.''} \emph{Journal of Machine Learning Research} 13 (January): 27--66.

\leavevmode\vadjust pre{\hypertarget{ref-cerda_2018}{}}%
Cerda, Patricio, Gaël Varoquaux, and Balázs Kégl. 2018. {``Similarity Encoding for Learning with Dirty Categorical Variables.''} \emph{Machine Learning} 107 (8): 1477--94. \url{https://doi.org/10.1007/s10994-018-5724-2}.

\leavevmode\vadjust pre{\hypertarget{ref-cerda_2020}{}}%
Cerda, P., and G. Varoquaux. 2020. {``Encoding High-Cardinality String Categorical Variables.''} \emph{IEEE Transactions on Knowledge and Data Engineering}, 1--1. \url{https://doi.org/10.1109/TKDE.2020.2992529}.

\leavevmode\vadjust pre{\hypertarget{ref-chambers_1992}{}}%
Chambers, JM, and TJ Hastie. 1992. {``Statistical Models. Chapter 2 of Statistical Models in s (1st Ed.).''} \emph{Routledge}. https://doi.org/\url{https://doi.org/10.1201/9780203738535}.

\leavevmode\vadjust pre{\hypertarget{ref-chen_2018}{}}%
Chen, Tianqi, Tong He, Michael Benesty, Vadim Khotilovich, Yuan Tang, Hyunsu Cho, Kailong Chen, et al. 2018. \emph{Xgboost: Extreme Gradient Boosting}. \url{https://CRAN.R-project.org/package=xgboost}.

\leavevmode\vadjust pre{\hypertarget{ref-chiquet_2016}{}}%
Chiquet, Julien, Yves Grandvalet, and Guillem Rigaill. 2016. {``On Coding Effects in Regularized Categorical Regression.''} \emph{Statistical Modelling} 16 (3): 228--37. \url{https://doi.org/10.1177/1471082X16644998}.

\leavevmode\vadjust pre{\hypertarget{ref-coors_2018}{}}%
Coors, Stefan. 2018. {``Automatic Gradient Boosting.''} Master's thesis, LMU Munich. \url{https://epub.ub.uni-muenchen.de/59108/1/MA_Coors.pdf}.

\leavevmode\vadjust pre{\hypertarget{ref-deleeuw_1976}{}}%
De Leeuw, Jan, Forrest W Young, and Yoshio Takane. 1976. {``Additive Structure in Qualitative Data: An Alternating Least Squares Method with Optimal Scaling Features.''} \emph{Psychometrika} 41 (4): 471--503.

\leavevmode\vadjust pre{\hypertarget{ref-dehghani2021benchmark}{}}%
Dehghani, Mostafa, Yi Tay, Alexey A. Gritsenko, Zhe Zhao, Neil Houlsby, Fernando Diaz, Donald Metzler, and Oriol Vinyals. 2021. {``The Benchmark Lottery.''} \url{https://arxiv.org/abs/2107.07002}.

\leavevmode\vadjust pre{\hypertarget{ref-delgado}{}}%
Fernández-Delgado, Manuel, Eva Cernadas, Senén Barro, and Dinani Amorim. 2014. {``Do We Need Hundreds of Classifiers to Solve Real World Classification Problems?''} \emph{Journal of Machine Learning Research} 15 (90): 3133--81. \url{http://jmlr.org/papers/v15/delgado14a.html}.

\leavevmode\vadjust pre{\hypertarget{ref-ferri_2009}{}}%
Ferri, C., J. Hernández-Orallo, and R. Modroiu. 2009. {``An Experimental Comparison of Performance Measures for Classification.''} \emph{Pattern Recognition Letters} 30 (1): 27--38. \url{https://doi.org/10.1016/j.patrec.2008.08.010}.

\leavevmode\vadjust pre{\hypertarget{ref-feurer_2015}{}}%
Feurer, Matthias, Aaron Klein, Katharina Eggensperger, Jost Springenberg, Manuel Blum, and Frank Hutter. 2015. {``Efficient and Robust Automated Machine Learning.''} In \emph{Advances in Neural Information Processing Systems 28}, edited by C. Cortes, N. D. Lawrence, D. D. Lee, M. Sugiyama, and R. Garnett, 2962--70. Curran Associates, Inc. \url{http://papers.nips.cc/paper/5872-efficient-and-robust-automated-machine-learning.pdf}.

\leavevmode\vadjust pre{\hypertarget{ref-friedman_2010}{}}%
Friedman, Jerome, Trevor Hastie, and Robert Tibshirani. 2010. {``Regularization Paths for Generalized Linear Models via Coordinate Descent.''} \emph{Journal of Statistical Software} 33 (1): 1--22. \url{https://doi.org/10.18637/jss.v033.i01}.

\leavevmode\vadjust pre{\hypertarget{ref-gelman_2006}{}}%
Gelman, Andrew, and Jennifer Hill. 2006. \emph{Data Analysis Using Regression and Multilevel/Hierarchical Models}. Cambridge university press.

\leavevmode\vadjust pre{\hypertarget{ref-grabczewski_2003}{}}%
Grąbczewski, Krzysztof, and Norbert Jankowski. 2003. {``Transformations of Symbolic Data for Continuous Data Oriented Models.''} In \emph{Artificial Neural Networks and Neural Information Processing --- ICANN/ICONIP 2003}, edited by Okyay Kaynak, Ethem Alpaydin, Erkki Oja, and Lei Xu, 359--66. Berlin, Heidelberg: Springer Berlin Heidelberg.

\leavevmode\vadjust pre{\hypertarget{ref-guo_2016}{}}%
Guo, Cheng, and Felix Berkhahn. 2016. {``Entity Embeddings of Categorical Variables.''} \emph{arXiv Preprint arXiv:1604.06737}.

\leavevmode\vadjust pre{\hypertarget{ref-hancock}{}}%
Hancock, John T, and Taghi M Khoshgoftaar. 2020. {``Survey on Categorical Data for Neural Networks.''} \emph{Journal of Big Data} 7: 1--41. \url{https://doi.org/10.1186/s40537-020-00305-w}.

\leavevmode\vadjust pre{\hypertarget{ref-hand_1997}{}}%
Hand, D. J., and W. E. Henley. 1997. {``Statistical Classification Methods in Consumer Credit Scoring: A Review.''} \emph{Journal of the Royal Statistical Society: Series A (Statistics in Society)} 160 (3): 523--41. \url{https://doi.org/10.1111/j.1467-985X.1997.00078.x}.

\leavevmode\vadjust pre{\hypertarget{ref-hornik_2007}{}}%
Hornik, Kurt, and David Meyer. 2007. {``Deriving Consensus Rankings from Benchmarking Experiments.''} In \emph{Advances in Data Analysis}, 163--70. Springer. \url{https://doi.org/10.1007/978-3-540-70981-7_19}.

\leavevmode\vadjust pre{\hypertarget{ref-hothorn_2005}{}}%
Hothorn, Torsten, Friedrich Leisch, Achim Zeileis, and Kurt Hornik. 2005. {``The Design and Analysis of Benchmark Experiments.''} \emph{Journal of Computational and Graphical Statistics} 14 (3): 675--99. \url{https://doi.org/10.1198/106186005X59630}.

\leavevmode\vadjust pre{\hypertarget{ref-kuhn_2019}{}}%
Kuhn, Max, and Kjell Johnson. 2019. \emph{Feature Engineering and Selection: A Practical Approach for Predictive Models}. Chapman; Hall/CRC. \url{http://www.feat.engineering/index.html}.

\leavevmode\vadjust pre{\hypertarget{ref-lang_2017}{}}%
Lang, Michel, Bernd Bischl, and Dirk Surmann. 2017. {``Batchtools: Tools for r to Work on Batch Systems.''} \emph{The Journal of Open Source Software} 2 (10). \url{https://doi.org/10.21105/joss.00135}.

\leavevmode\vadjust pre{\hypertarget{ref-mair_2010}{}}%
Mair, Patrick, and Jan de Leeuw. 2010. {``A General Framework for Multivariate Analysis with Optimal Scaling: The r Package Aspect.''} \emph{Journal of Statistical Software} 32 (1): 1--23. \url{https://doi.org/10.18637/jss.v032.i09}.

\leavevmode\vadjust pre{\hypertarget{ref-meyer_2018}{}}%
Meyer, David, and Kurt Hornik. 2018. \emph{Relations: Data Structures and Algorithms for Relations}. \url{https://CRAN.R-project.org/package=relations}.

\leavevmode\vadjust pre{\hypertarget{ref-micci_barreca_2001}{}}%
Micci-Barreca, Daniele. 2001. {``A Preprocessing Scheme for High-Cardinality Categorical Attributes in Classification and Prediction Problems.''} \emph{SIGKDD Explor. Newsl.} 3 (1): 27--32. \url{https://doi.org/10.1145/507533.507538}.

\leavevmode\vadjust pre{\hypertarget{ref-nadeau_2003}{}}%
Nadeau, Claude, and Yoshua Bengio. 2003. {``Inference for the Generalization Error.''} \emph{Machine Learning} 52 (3): 239--81. \url{https://doi.org/10.1023/A:1024068626366}.

\leavevmode\vadjust pre{\hypertarget{ref-niesl2021overoptimism}{}}%
Nießl, Christina, Moritz Herrmann, Chiara Wiedemann, Giuseppe Casalicchio, and Anne-Laure Boulesteix. n.d. {``Over-Optimism in Benchmark Studies and the Multiplicity of Design and Analysis Options When Interpreting Their Results.''} \emph{WIREs Data Mining and Knowledge Discovery} n/a (n/a): e1441. https://doi.org/\url{https://doi.org/10.1002/widm.1441}.

\leavevmode\vadjust pre{\hypertarget{ref-probst_2019}{}}%
Probst, Philipp, Marvin N. Wright, and Anne-Laure Boulesteix. 2019. {``Hyperparameters and Tuning Strategies for Random Forest.''} \emph{Wiley Interdisciplinary Reviews: Data Mining and Knowledge Discovery} 0 (0): e1301. \url{https://doi.org/10.1002/widm.1301}.

\leavevmode\vadjust pre{\hypertarget{ref-prokhorenkova_2018}{}}%
Prokhorenkova, Liudmila, Gleb Gusev, Aleksandr Vorobev, Anna Veronika Dorogush, and Andrey Gulin. 2018. {``CatBoost: Unbiased Boosting with Categorical Features.''} In \emph{Advances in Neural Information Processing Systems 31}, edited by S. Bengio, H. Wallach, H. Larochelle, K. Grauman, N. Cesa-Bianchi, and R. Garnett, 6638--48. Curran Associates, Inc. \url{http://papers.nips.cc/paper/7898-catboost-unbiased-boosting-with-categorical-features.pdf}.

\leavevmode\vadjust pre{\hypertarget{ref-prokopev_2018}{}}%
Prokopev, Viacheslav. 2018. {``Mean (Likelihood) Encodings: A Comprehensive Study.''} \emph{Kaggle Forums}. Kaggle. \url{https://www.kaggle.com/vprokopev/mean-likelihood-encodings-a-comprehensive-study}.

\leavevmode\vadjust pre{\hypertarget{ref-r_2021}{}}%
R Core Team. 2021. \emph{R: A Language and Environment for Statistical Computing}. Vienna, Austria: R Foundation for Statistical Computing. \url{https://www.R-project.org/}.

\leavevmode\vadjust pre{\hypertarget{ref-rodriguez_2018}{}}%
Rodríguez, Pau, Miguel A. Bautista, Jordi Gonzàlez, and Sergio Escalera. 2018. {``Beyond One-Hot Encoding: Lower Dimensional Target Embedding.''} \emph{Image and Vision Computing} 75: 21--31. https://doi.org/\url{https://doi.org/10.1016/j.imavis.2018.04.004}.

\leavevmode\vadjust pre{\hypertarget{ref-schliep_2016}{}}%
Schliep, Klaus, and Klaus Hechenbichler. 2016. \emph{Kknn: Weighted k-Nearest Neighbors}. \url{https://CRAN.R-project.org/package=kknn}.

\leavevmode\vadjust pre{\hypertarget{ref-seca_2021}{}}%
Seca, Diogo, and João Mendes-Moreira. 2021. {``Benchmark of Encoders of Nominal Features for Regression.''} In \emph{Trends and Applications in Information Systems and Technologies}, edited by Álvaro Rocha, Hojjat Adeli, Gintautas Dzemyda, Fernando Moreira, and Ana Maria Ramalho Correia, 146--55. Cham: Springer International Publishing.

\leavevmode\vadjust pre{\hypertarget{ref-steinwart_2017}{}}%
Steinwart, Ingo, and Philipp Thomann. 2017. {``{liquidSVM}: A Fast and Versatile SVM Package.''} \emph{{ArXiv e-Prints 1702.06899}}, February. \url{http://www.isa.uni-stuttgart.de/software}.

\leavevmode\vadjust pre{\hypertarget{ref-therneau_2018}{}}%
Therneau, Terry, and Beth Atkinson. 2018. \emph{Rpart: Recursive Partitioning and Regression Trees}. \url{https://CRAN.R-project.org/package=rpart}.

\leavevmode\vadjust pre{\hypertarget{ref-thomas_2018}{}}%
Thomas, Janek, Stefan Coors, and Bernd Bischl. 2018. \emph{Automatic Gradient Boosting}. \url{http://arxiv.org/abs/1807.03873v2}.

\leavevmode\vadjust pre{\hypertarget{ref-thornton_2013}{}}%
Thornton, Chris, Frank Hutter, Holger H. Hoos, and Kevin Leyton-Brown. 2013. {``Auto-WEKA: Combined Selection and Hyperparameter Optimization of Classification Algorithms.''} In \emph{Proceedings of the 19th ACM SIGKDD International Conference on Knowledge Discovery and Data Mining}, 847--55. KDD '13. New York, NY, USA: ACM. \url{https://doi.org/10.1145/2487575.2487629}.

\leavevmode\vadjust pre{\hypertarget{ref-tutz_2016}{}}%
Tutz, Gerhard, and Jan Gertheiss. 2016. {``Rejoinder: Regularized Regression for Categorical Data.''} \emph{Statistical Modelling} 16 (3): 249--60. \url{https://doi.org/10.1177/1471082X16652780}.

\leavevmode\vadjust pre{\hypertarget{ref-vanschoren_2013}{}}%
Vanschoren, Joaquin, Jan N. van Rijn, Bernd Bischl, and Luis Torgo. 2013. {``{OpenML}: Networked Science in Machine Learning.''} \emph{SIGKDD Explorations} 15: 49--60. \url{https://doi.org/10.1145/2641190.2641198}.

\leavevmode\vadjust pre{\hypertarget{ref-weinberger_2009}{}}%
Weinberger, Kilian, Anirban Dasgupta, John Langford, Alex Smola, and Josh Attenberg. 2009. {``Feature Hashing for Large Scale Multitask Learning.''} In \emph{Proceedings of the 26th Annual International Conference on Machine Learning}, 1113--20. ICML '09. New York, NY, USA: Association for Computing Machinery. \url{https://doi.org/10.1145/1553374.1553516}.

\leavevmode\vadjust pre{\hypertarget{ref-wright_2019}{}}%
Wright, Marvin N., and Inke R. König. 2019. {``Splitting on Categorical Predictors in Random Forests.''} \emph{PeerJ} 7. \url{https://doi.org/10.7717/peerj.6339}.

\leavevmode\vadjust pre{\hypertarget{ref-wright_2017}{}}%
Wright, Marvin N., and Andreas Ziegler. 2017. {``{ranger}: A Fast Implementation of Random Forests for High Dimensional Data in {C++} and {R}.''} \emph{Journal of Statistical Software} 77 (1): 1--17. \url{https://doi.org/10.18637/jss.v077.i01}.

\leavevmode\vadjust pre{\hypertarget{ref-young_1976}{}}%
Young, Forrest W, Jan De Leeuw, and Yoshio Takane. 1976. {``Regression with Qualitative and Quantitative Variables: An Alternating Least Squares Method with Optimal Scaling Features.''} \emph{Psychometrika} 41 (4): 505--29. https://doi.org/\url{https://doi.org/10.1007/BF02296972}.

\end{CSLReferences}

\bibliographystyle{spbasic}
\bibliography{literature.bib}

\end{document}